\documentclass[10pt,journal,compsoc]{IEEEtran}

\usepackage[nocompress]{cite}
\usepackage{multirow}
\usepackage{import}
\usepackage{times}
\usepackage{breakcites}

\usepackage[usenames,dvipsnames]{xcolor}

\usepackage{makeidx}
\usepackage{color}
\usepackage{cite}
\usepackage{algorithm}
\usepackage{algorithmic}
\usepackage{blkarray}
\usepackage{url}
\usepackage{enumitem}

\usepackage{booktabs}
\usepackage{multirow}
\usepackage{array}
\usepackage{tabularx}

\usepackage{pbox}
\usepackage{wrapfig}
\usepackage{times}
\usepackage{epsfig}
\usepackage{graphicx}
\usepackage{amsmath}
\usepackage{amssymb}
\usepackage{url}

\newcommand{\iou}{IoU}

\usepackage[usenames,dvipsnames]{xcolor}

\usepackage{booktabs}
\usepackage{array}

\usepackage{pbox}
\usepackage{wrapfig}
\usepackage{subfigure}

\usepackage[color]{changebar}
 \cbcolor{red}
\newcommand{\addd}{\begin{changebar}\cbcolor{blue}}

\usepackage{stfloats}

\usepackage{url}

\begin{document}
\title{Efficient 2D and 3D Facade Segmentation using Auto-Context}

\author{Raghudeep Gadde*,
        Varun Jampani*,
        Renaud Marlet,
        and~Peter~V.~Gehler
        \IEEEcompsocitemizethanks{
            \IEEEcompsocthanksitem Raghudeep Gadde and Renaud Marlet are with Universit\'e Paris-Est, LIGM (UMR 8049), CNRS, ENPC, ESIEE Paris, UPEM, France.
            \IEEEcompsocthanksitem Varun Jampani is with MPI-IS, Germany.
            \IEEEcompsocthanksitem Peter Gehler is with BCCN, University of Tubingen and MPI-IS, Germany.
            \IEEEcompsocthanksitem The first two authors have contributed equally to this work.}}

\IEEEtitleabstractindextext{%
\begin{abstract}
This paper introduces a fast and efficient segmentation technique
for 2D images and 3D point clouds of building facades.
Facades of buildings are highly structured and
consequently most methods that have been proposed for this problem
aim to make use of this strong prior information. Contrary to most prior work,
we are describing
a system that is almost domain independent and consists of standard
segmentation methods. We train a sequence of boosted decision trees using
auto-context features. This is learned using
stacked generalization. We find that this
technique performs better, or comparable with all previous published methods and
present empirical results on \textit{all} available 2D and 3D facade benchmark datasets.
The proposed method is simple to implement, easy to extend, and very
efficient at test-time inference.

\end{abstract}

}

\maketitle

\IEEEdisplaynontitleabstractindextext

\IEEEpeerreviewmaketitle

\IEEEraisesectionheading{\section{Introduction}\label{sec:introduction}}

\begin{figure*}[t]
\setlength\abovecaptionskip{-1pt}
\includegraphics[width=\textwidth]{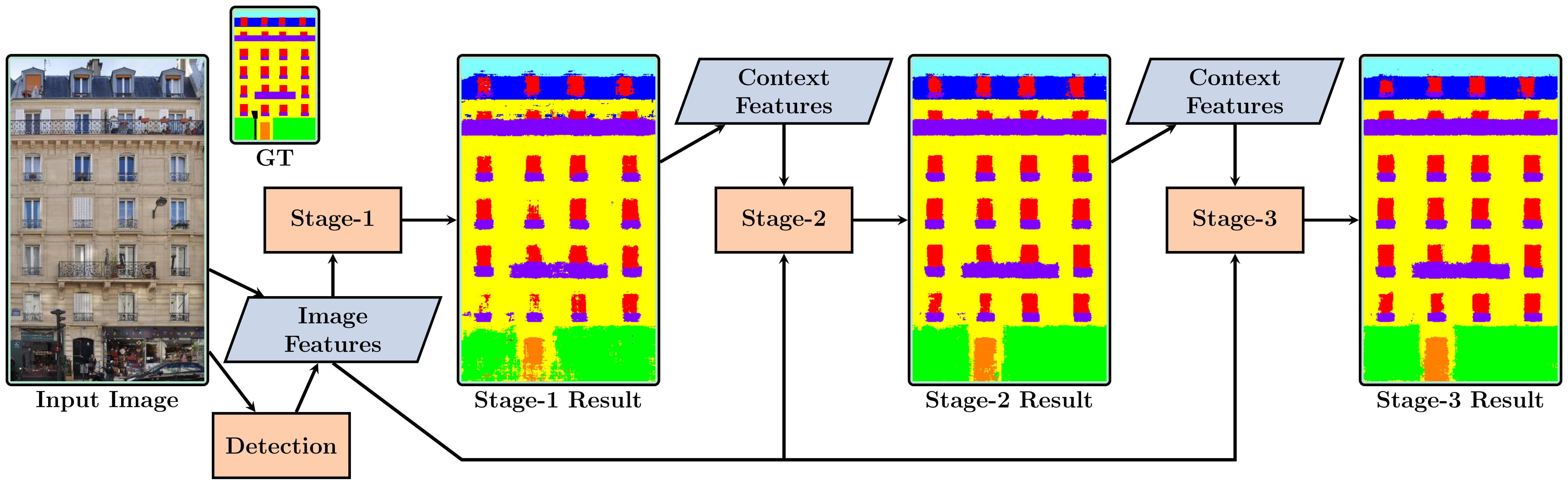}
\caption{Schematic of different components in our facade segmentation pipeline with a
sample facade from ECP dataset~\cite{teboul2010ecole}. `Image' and `Context' features
correspond to features extracted on input image and previous-stage segmentation result
respectively. `Stage-$n$' refers to the $n^{th}$ stage auto-context classifier.
The segmentation result is successively refined by the auto-context classifiers from
their respective previous stage result.}
\label{fig:overview}
\end{figure*}

\IEEEPARstart{I}n this paper, we consider the problem of segmenting building facades
in an image, resp. a point cloud, into different
semantic classes. An example
image from a common benchmark dataset for this problem is shown in
Fig.~\ref{fig:overview} along with a manual annotation.
Being able to segment facades is a core component of several real world applications 
in urban modeling, such as thermal performance evaluation and shadow casting on windows.
As evident from the example in Fig.~\ref{fig:overview},
images of buildings exhibit a strong
structural organization due to architectural design choices and construction
constraints. For example, windows are usually not placed randomly, but
on the same height; a door can only be found on the street-level, etc.

This problem is also an interesting test-bed for general-purpose
segmentation methods that also allow strong architectural prior. 
As a result, it appears
reasonable to assume that methods which incorporate such high-level
knowledge will perform well in doing automatic facade
segmentation. Following this, existing facade segmentation methods use
complex models and inference technique to incorporate high-level
architectural knowledge for better pixel-level segmentation.
Some examples are Conditional Random Field (CRF) models
that use higher-order potential
functions~\cite{yang2011hierarchical,Tylecek13}. Another route are
grammar-based models that include generative
rules~\cite{riemenschneider2012irregular,teboul2011rl,martinovic2013bayesian}
and try to infer the choice of production rules at parse time from the image evidence.

Contrary to the philosophy of existing methods, we largely ignore domain-specific
knowledge. We describe a generic segmentation method that is
easy to implement, has fast test-time inference, and is easily
adaptable to new datasets. Our key observation is that very good
segmentation results can be achieved by pixel classifications methods that
use basic image features in conjunction with auto-context
features~\cite{tu2008auto}. In
this work, we develop a simple and generic auto-context-based
framework for facade segmentation. The system is a sequence of boosted
decision tree classifiers, that are stacked using
auto-context~\cite{tu2008auto} features and learned using stacked
generalization~\cite{Wolpert92stackedgeneralization}. We stack three pixel
classifiers using auto-context features for images and two
classifiers for 3D point clouds. Fig.~\ref{fig:overview}
shows an example segmentation result for various classification stages of our
method. As can be seen in the visual result of Fig. 1, the segmentation
result is successively refined by the auto-context classifiers, from their
respective previous stage result.
Using pixel-level classifiers along with generic image
features has the advantage of being versatile and fast compared to
existing complex methods. The entire
pipeline consists of established components and we consider it to be a
baseline method for this task. Surprisingly, our auto-context based method,
despite being simple and generic, consistently performs better or on par with
existing complex methods on all the available
diverse facade benchmark datasets in both 2D and 3D. Moreover, the presented
approach has favourable runtime in comparison to existing
approaches for facade segmentation. Therefore, this approach defines a new
state-of-the-art method in terms of empirical performance.
It is important to note that by proclaiming so, we are not invalidating the use of
existing methods, that make of domain-knowledge, for facade segmentation.
Experiments suggest that more domain-specific models would benefit
from better unary predictions from our approach.
Moreover our findings also suggest
that previous methods need to be carefully re-evaluated in terms of a
relative improvement compared to a method like the proposed one.

A pixel or point-wise facade classification might not be a desired output for some applications.
For instance, high level structural information is needed 
to construct Building Information Models (BIMs). 
We show how the pixel predictions we obtain can be used in a inverse procedural modeling
system~\cite{teboul2011rl} that 
parses facade images. These rules are of interest in different applications
and we show that an improved pixel-wise predictions directly translates
into a better facade parsing result.

This paper is organized as follows. Related work is discussed in
Section~\ref{sec:related}, followed by a detailed description of the
auto-context segmentation setup in Section~\ref{sec:method}.
Section~\ref{sec:experiment} contains the experimental results and 
we conclude in Section~\ref{sec:conclusion}.

\section{Related Work}
\label{sec:related}

Facade segmentation approaches can be broadly classified into two
categories: \textit{bottom-up
  methods}~\cite{martinovic2012three,Tylecek13,yang2011hierarchical,cohen2014dp}
that use pixel-level classifiers in combination with CRF models and
\textit{top-down
  methods}~\cite{teboul2011rl,riemenschneider2012irregular,martinovic2013bayesian,kozinski2014beyond,kozinski2015mrfshapeprior}
that use shape grammars or a user defined shape prior.
The shape grammar methods seek to parse a facade in terms of
a set of production rules and element attributes, thus
segmenting the facade into semantic regions. The central idea
is to represent the facade using a parse tree and search for the
grammar derivation that best matches a pixel-level classification
of an image.
The high structural
organization of facades due to architectural design choices make such
a generative approach a natural model candidate. However it is 
not easily amendable to efficient inference, which
often leads to inefficient and sub-optimal segmentation results.
Furthermore, due to the strong prior that a grammar imposes, they are
not necessarily pixel-wise accurate. As a
consequence, the state-of-the-art methods in terms of pixel accuracy
are dominated by the bottom-up methods, although they do not provide
structured information as in a parse tree.

In~\cite{martinovic2012three,mathias2015atlas}, a three-layered system is proposed.
A first layer uses a recursive neural network to obtain pixel label
probabilities, which are fed into a grid CRF model in a second layer
along with object detections. The third layer enforces weak
architectural principles in facades as post-processing. This setup
combines high-level and low-level information into a single
prediction. The runtime of this system is mentioned
in~\cite{cohen2014dp} to be about 2
minutes for an image of size 500 by 300. Other
approaches~\cite{yang2011hierarchical,Tylecek13} incorporate architectural
knowledge in a single CRF framework using higher-order potential
functions. The method of~\cite{yang2011hierarchical} proposes a
hierarchical CRF framework to encode inter-class location information
in facades. The work of~\cite{Tylecek13} uses long-range pairwise and
ternary potentials to encode the repetitive nature of various class
regions. Both methods require specific inference techniques that result
in non-negligible runtimes. The approach of~\cite{cohen2014dp} is to
use a sequence of dynamic programming runs that search for optimal
placement of window and balcony rows, door location and others. Every
single step is very fast and the overall system is mostly global
optimal. The downside is that the sequence and type of
classifications needs to match facade architecture type.
\cite{kozinski2015mrfshapeprior} employ a user-defined shape prior (an adjacency pattern)
for parsing rectified facade images 
and formulates parsing as a MAP-MRF problem over a pixel grid.

Recently, techniques have been introduced for facade understanding and
modeling in 3D~\cite{riemenschneider2014learning,martinovic20153d}.
The 3D point cloud or meshes that these methods operate on are
constructed using 2D images captured from multiple viewpoints.
A standard way to label a 3D mesh or a point cloud, is to label all the
overlapping images used for reconstructing the 3D model and then fuse
the 2D predictions to obtain a consistently
labeled 3D model~\cite{ladicky2010what,tighe2010superparsing}.
The work of~\cite{riemenschneider2014learning} proposed a fast
technique to segment 3D facade meshes by exploiting the geometry of the
reconstructed 3D model. To label a mesh face, their approach selects a single 2D image
(from the set of images used for reconstruction) that best captures the
semantics. The speed of this technique comes at the cost of performance.
The method of~\cite{martinovic20153d} implements a three-stage approach to label point clouds
of facades directly in 3D.
First, features on 3D points are computed and are classified into various semantic classes.
Next, facades belonging to different buildings
are separated based on previously obtained semantics. Finally,
weak architectural rules are applied to enforce structural priors,
leading to marginal improvements in performance (0.78\% \iou) compared to the inital
classifier predictions.

All the discussed methods build on top of semantic label probabilities
which are obtained using pixel/point classifiers. It is only after those have
been obtained that architectural constraints are taken into account.
In the system we describe in this paper,
2D or 3D segmentations are obtained only using image or point cloud 
auto-context features without resorting to any domain specific
architectural constraints. As a result, several above mentioned domain-specific
approaches would benefit from using the segmentation label probabilities
obtained with our proposed domain-independent approach.

The closest to our work are~\cite{frohlich2013semantic} and~\cite{gatta2014stacked},
which also proposed auto-context based
methods for facade segmentation.
\cite{frohlich2013semantic} incorporated auto-context features in
random decision forests where the classification results from
top-layers of trees are used to compute auto-context features and are
then used in training the lower layers of the trees in the forest. More
recently,~\cite{gatta2014stacked} proposed to use the local Taylor
coefficients computed from the posterior at different scales as
auto-context. Although~\cite{frohlich2013semantic} and~\cite{gatta2014stacked}
are conceptually similar, the method we propose uses
different low-level features, different auto-context features and different learning
techniques achieving better performance on benchmark datasets.

\section{Auto-Context Segmentation}
\label{sec:method}

We propose an architecture that combines standard segmentation methods into a single framework.
Boosted decision trees are stacked with
the use of auto-context~\cite{tu2008auto} features from the second
layer onward. This system is then trained using stacked
generalization~\cite{Wolpert92stackedgeneralization}. We will describe the ingredients in the
following, starting with the segmentation algorithm
(Sec.~\ref{sec:pixel-wise-segm}), the feature representation for images
(Sec.~\ref{sec:image-features}) and for point-clouds (Sec.~\ref{sec:pcl-features}), auto-context features
(Sec.~\ref{sec:autocontext-features}), and the training procedure
(Sec.~\ref{sec:stack-gener}).

Given a point cloud or an image $I$, the task of semantic segmentation is to classify every
point or pixel $i$ into one of $C$ classes $c_i\in\{1,\ldots,C\}$. During training,
we have access to a set of $N$ class-annotated images each with a variable number of points/pixels:
$(I_i^j,c_i^j), j=1,\ldots,N$.
We will comment on the
loss function in the experimental section and for now treat the
problem as one that decomposes over the set of pixels. Two different
feature sets are distinguished, data-dependent features
$f_i\in\mathbb{R}^{D_f}$ that are derived from the spatial and color observations
in a point cloud or image, and auto-context features $a_i\in\mathbb{R}^{D_a}$ based on
the prediction results from previous stages.

\vspace{-0.2cm}
\subsection{Model Architecture}\label{sec:pixel-wise-segm}

Our system consists of a sequence of classifiers as suggested
in~\cite{tu2008auto}. A schematic overview of the pipeline is depicted
in Fig.~\ref{fig:overview}. At every stage $t$, the classifier has access
to the image and to predictions of all earlier stages. Formally, at
stage $t>1$ and at each pixel $i$,
a classifier $F^t$ maps image ($I$) and auto-context features ($a_i$) to
a probability distribution $P^t$ of the pixel class assignments
\begin{equation}
  \label{eq:1}
  F^t\left(f_i(I), a_i(P^{t-1})\right) \mapsto P^t(c_i|I), \forall i.
\end{equation}
For pixel classifier $F^t$, we use boosted decision trees that store
conditional distributions at their leaf nodes. In general, the output
of $F^t$ need not be a distribution. The first stage $t=1$ depends only on the
features $F^1(f_i(I))$ derived directly from the image. The setting is identical
for point clouds.

This architecture is a conceptually easy and efficient way to use
contextual information in pixel-level classification. Classifiers of
later stages can correct errors that earlier stages make. An example
sequence of predictions can be seen in Fig.~\ref{fig:overview}.
For example, an auto-context feature can encode the density
of a predicted class around a pixel. The classifier can learn that
certain classes only appear in clusters which then allows to remove
spurious predictions. This has a similar smoothing effect as some
pairwise CRF models have, but with the benefit of a much faster
inference.

\subsection{Image Features}\label{sec:image-features}

As image features, we computed 17 TextonBoost filter
responses~\cite{shotton2006textonboost}, location information, RGB
color information, dense Histogram of Oriented
Gradients~\cite{dalal2005histograms}, Local Binary Pattern features~\cite{ojala2002multiresolution},
and all filter averages over image rows and columns at
each pixel. These are computed using the
DARWIN~\cite{Gould:JMLR2012} toolbox.

\begin{figure}
\setlength\abovecaptionskip{-1pt}
\centering
  \subfigure[Facade]{%
    \includegraphics[width=.3\columnwidth]{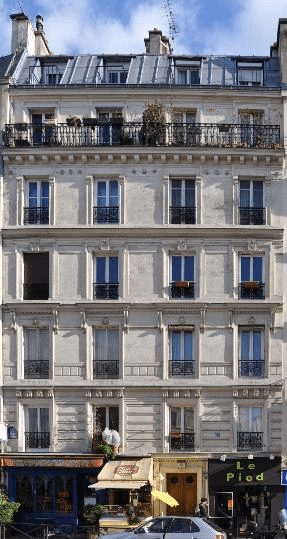} \label{fig:samplefacade}
  }
  ~~~~~~
  \subfigure[Detection]{%
    \includegraphics[width=.3\columnwidth]{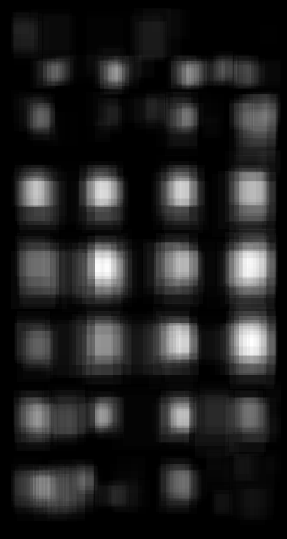} \label{fig:detection}
  }
  \caption{(a) A facade and (b) its window detection scores.
    Bright and dark regions correspond to high and low detection
    scores respectively.}
\label{fig:detection}
\vspace{-0.4cm}
\end{figure}

In addition to the above generic segmentation features, we include
detection scores for some specific objects.
Following~\cite{martinovic2012three,ok20123dimpvt,kozinski2014graphgram}, we use detectors for windows as well as
doors. Whereas~\cite{martinovic2012three,ok20123dimpvt} fused the detection scores
into the output of the pixel classifiers, we turned the detection
scores into image features at every single pixel. We use the integral
channel features detector from~\cite{dollar2009integral} for which a
toolbox is available~\cite{PMT}. For a given image, the
detector outputs a number of bounding boxes along with a corresponding
score for each bounding box. We sum up the scores to get a single
detection score at each pixel. Object detection parameters are automatically
estimated using the training data to get a good recall.
Fig.~\ref{fig:detection} shows an example window
detection output for a sample facade image. The detection feature is
of course a problem dependent one and based on the prior knowledge
about the special classes: door and windows. However it is still a
generic feature in the sense that the prior information is extremely weak
and a generic detection system has been used to obtain it. Moreover,
door and window classes are common to any architecture of facades. In the end, 761
low-level image features coupled with 2 door and window detection
features make a total of 763 feature values at each pixel.

\subsection{Point Cloud Features}\label{sec:pcl-features}
We use the same set of features as~\cite{martinovic20153d} to describe a
point cloud. The features inlude the mean RGB color values, their corresponding
LAB values, the estimated normal at the 3D point, the spin image descriptor~\cite{johnson1999using},
the height of a
point above an estimated groud plane, the depth of the point from an estimated
facade plane and the inverse height of the point which is the distance from the
uppermost point of the facade in the direction of the gravity vector. The combination of all these
features form a 132-dimensional vector for every point.

\subsection{Auto-context Features}\label{sec:autocontext-features}

In addition to the image features, the classifiers from stage $t>1$
can condition on statistics computed from previous predictions. We
include the auto-context features $a_i$ that are computed from
predictions of the previous classifier $P^{t-1}(\cdot | I)$ only. For
every pixel $i$ we compute the following auto-context
features of length $14C+1$, where $C$ is the number of classes.

\begin{itemize}[label={},leftmargin=*]
\setlength\itemsep{-1em}
\item \textbf{Class probability:}~  The probability $P^{t-1}(c_i|I)$.
    \small(length $C$). \normalsize\\
\item \textbf{Entropy:}~  The entropy of $P^{t-1}(\cdot|I)$. This
feature quantifies the ambiguity of the $t-1$ stage prediction \small (length 1)\normalsize.\\
\item \textbf{Row and column scores:}~ We compute the percentage of
predicted classes in the row and column of pixel $i$. Along with this
percentage, we compute the average score of all pixels in the same row
and column as $i$ \small (length $4C$)\normalsize.\\
\item \textbf{Distance to the nearest class:}~  Both Euclidean and
Manhattan distances to the nearest class pixel are computed as
features \small (length $2C$)\normalsize.\\
\item \textbf{Class color model:}~  For every class $c$ we fit, with maximum
likelihood, a Gaussian distribution to the RGB values of all
pixels that are being predicted to be of class $c$. To be more robust, we
fit the distribution only to  pixels with probabilities greater
than the $3{rd}$ quartile. For every pixel, we then calculate the
log-likelihood for all classes \small (length $C$)\normalsize.\\
\item \textbf{Bounding box features:}~  For every class, we fit a rectangular
bounding box to every connected component of MAP predictions. For
every pixel we compute a $C$ dimensional vector with the $c$'th component being a
$1$ or $0$ depending on whether it lies inside or outside of a box for
class $c$. A variant of this feature is to compute the average class
probability inside the box. This feature aims to improve the
segmentation of rectangular objects such as doors and windows \small (length
$2C$)\normalsize.\\
\item \textbf{Neighborhood statistics:}~  For every pixel, the average class
probability is computed in a $10\times5$ region above and below the
pixel; and also in a $5\times10$ region left and right to that pixel \small (length $4C$)\normalsize.\\
\end{itemize}
\vspace{-0.3cm}

In the case of point clouds, we use only the class probabilities and the
entropy of the class probabilities as auto-context features. So, for point
clouds the size of the auto-context features is $C+1$.

\subsection{Stacked Generalization}\label{sec:stack-gener}

We train the sequence of classifiers using stacked
generalization~\cite{Wolpert92stackedgeneralization}. The training
data is split in $M$ folds and at each stage, $M$ different models are trained
using data from $M-1$ folds, with one fold held out. The $M$ models are used to
obtain prediction on the held out fold, this results in a set of
cross-validation predictions. It is from these predictions that the
auto-context features for training are computed. The next stage
classifier is trained subsequently, in the same manner. For every
stage, one additional classifier is trained using the entire training data (all
$M$ folds) that is used during test-time inference.
In our experiments, to segment 2D images we divide the training set
into four folds ($M=4$) and for 3D point clouds, we do not use the stacked generalization
($M=1$) due to the availability of fewer training points.
We use three classification stages for 2D images and
only two classification stages for 3D point clouds, as we observe that
the performance levels out after that.

Thus, instead of using single classifier in each stage,
the auto-context features are computed using predictions from
different classifiers, different also from the classifier that will be
used at test time. The reason for this procedure is to obtain
features that are not computed on training predictions
and thus avoid to overfit to the data. This procedure is a standard
strategy and is found to be stable and well performing in many scenarios, e.g.\cite{gehler2009feature}.

For training and testing, we used the DARWIN
toolbox~\cite{Gould:JMLR2012}. The maximum tree-depth of each
boosted decision tree classifier is set to two and we used a maximum of 200 boosting rounds.

\section{Experiments}
\label{sec:experiment}

We evaluate the auto-context pipeline on all seven benchmark datasets
that are available for the problem of facade segmentation. For all
datasets except LabelMeFacade~\cite{frohlich2010fast} and
RueMonge2014~\cite{riemenschneider2014learning} datasets, we report five fold
cross-validation results, the standard protocol used in the
literature. One fold cross-validation is done for LabelMeFacade and RueMonge2014
datasets as the train and test data splits are pre-specified for these
datasets. We compare against all recent best performing methods.

As performance measures, we use the overall pixel-wise classification
accuracy, the accuracy averaged over the classes and the
intersection over union (\iou) score, popularized by the VOC segmentation
challenges~\cite{everingham2010pascal}.
The \iou~score is a higher-order loss function and Bayes optimal
prediction requires dedicated inference techniques. For simplicity, we
report MAP predictions for all pixels and evaluate all three measures
on this prediction as done in the literature concerning these datasets.
 The three measures are defined as follows in terms of false positives
(FP), true positives (TP), and false negatives (FN).
\begin{itemize}
\item \textit{Overall Pixel Accuracy}: ``TP / (TP + FN)" computed over
  entire image pixels of all classes.
\item \textit{Average Class Accuracy}: Pixel accuracy computed for all
  classes separately and then averaged.
\item \textit{Intersection Over Union Score} (\iou): ``TP / (TP + FN +
  FP)'' computed on every class and then averaged.
\end{itemize}

The performance differences are tested for statistical significance.
We used a paired t-test with one tail and $p<0.01$.

\subsection{Datasets}
\label{sec:datasets}
\noindent\textbf{ECP Dataset.} The ECP dataset~\cite{teboul2010ecole}
consists of 104 rectified facade images of Hausmannian architectural buildings
from Paris. For five-fold cross validation, we randomly divide the training data
into 4 sets of 20 images and 1 set of 24 images as in~\cite{martinovic2012three}. There are seven
semantic classes in this dataset.

\noindent\textbf{Graz Dataset.} This
dataset~\cite{riemenschneider2012irregular} has 50 facade images of
various architectures (Classicism, Biedermeier, Historicism, Art
Nouveau) from buildings in Graz. There are only four semantic classes,
and the data is divided into 5 equal sets for cross-validation.

\noindent\textbf{eTRIMS Dataset.} The eTRIMS
dataset~\cite{korc-forstner-tr09-etrims} consists of 60 non-rectified
images. Facades in this dataset are more
irregular and follow only weak architectural principles. Again, we split the data
into 5 equal sets for cross-validation.

\noindent\textbf{CMP Dataset.} This dataset, proposed
in~\cite{Tylecek13}, has 378 rectified facades of diverse styles and 12 semantic classes
 in its base set. We divided the data into 4 sets of 75 images each and one set of 78
images for cross-validation.

\noindent\textbf{LabelMeFacade Dataset.} Introduced in~\cite{frohlich2010fast}, this
 dataset has 100 training and 845 testing facade images taken from LabelMe segmentation
 dataset~\cite{russell2008labelme}. Facades in this dataset are highly
 irregular with a lot of diversity across images.

\noindent\textbf{ENPC Art-deco dataset.} This dataset, first used in~\cite{gadde2016learninggrammars},
contains 79 rectified and cropped facade images of the Art-deco style
buildings from Paris. Similar to the ECP dataset, the images in this
dataset are segmented into seven semantic classes.

\noindent\textbf{RueMonge2014 Dataset.} This dataset, introduced in~\cite{riemenschneider2014learning},
is aimed towards providing a benchmark for 2D and 3D facade segmentation,
and inverse procedural modeling. It consists of 428 high-resolution and multi-view
images of facades following the Haussmanian style architecture, a reconstructed
point cloud, a reconstructed mesh and a framework to evaluate segmentation results.
Three tasks are proposed on this dataset in~\cite{riemenschneider2014learning,martinovic20153d}
\footnote{\url{http://varcity.eu/3dchallenge/}}.
The first task is the standard image labeling task where each pixel has
to be assigned a semantic label. The second task is the mesh labeling task where
a semantic label has to be assigned to each face of a given mesh. And the third task
is the point cloud labeling task where a semantic label has to be assigned to
each point in the point cloud. For each of the tasks, fixed splits in training
and testing sets are pre-defined. The ground-truth labeling consists of
seven semantic classes, same as in the ECP dataset.

\begin{figure}[t]
  \scriptsize
  \centering
    \begin{tabular}{p{0.02cm}p{0.95cm}|>{\centering\arraybackslash}p{0.25cm}>
{\centering\arraybackslash}p{0.25cm}>{\centering\arraybackslash}p{0.25cm}>
{\centering\arraybackslash}p{0.25cm}>{\centering\arraybackslash}p{0.25cm}>
{\centering\arraybackslash}p{0.25cm}>{\centering\arraybackslash}p{0.3cm}|>
{\centering\arraybackslash}p{0.25cm}>{\centering\arraybackslash}p{0.25cm}>
{\centering\arraybackslash}p{0.25cm}}
      \toprule
      & &\rotatebox[origin=c]{45}{Door} & \rotatebox[origin=c]{45}{Shop} & \rotatebox[origin=c]{45}{Balcony} & \rotatebox[origin=c]{45}{Window} & \rotatebox[origin=c]{45}{Wall} & \rotatebox[origin=c]{45}{Sky} & \rotatebox[origin=c]{45}{Roof} & \rotatebox[origin=c]{45}{\textbf{Average}} & \rotatebox[origin=c]{45}{\textbf{Overall}} & \rotatebox[origin=c]{45}{\textbf{IoU}}\\
      \midrule
      & \textbf{~\cite{martinovic2012three}} & 60 & 86 & 71 & 69 & \textbf{93} & 97 & 73 & 78.4 & 85.1 & -\\[0.1cm]
      & \textbf{~\cite{mathias2015atlas}} & 58 & 97 & 81 & 76 & 90 & 94 & 87 & 83.4 & 88.1 & - \\[0.1cm]
      & \textbf{~\cite{cohen2014dp}} & 79 & 94 & 91 & 85 & 90 & 97 & 90 & 89.4 & 90.8 & -\\[0.1cm]
      & \textbf{\cite{kozinski2015mrfshapeprior}} & 79 & \textbf{97} & 91 & \textbf{87} & 90 & 97 & 91 & 90.3 & \textbf{91.3} & - \\[0.1cm]
      & \textbf{ST1} & 76 & 88 & 86 & 77 & 92 & 97 & 87 & 86.0 & 88.9 & 75.3\\[0.1cm]
      & \textbf{ST2} & 79 & 90 & 89 & 81 & 92 & \textbf{98} & 88 & 88.3 & \textbf{90.5} & 78.6 \\[0.1cm]
      & \textbf{ST3} & 80 & 92 & 89 & 82 & 92 & \textbf{98} & 88 & 88.8 & \textbf{90.8} & 79.3 \\[0.1cm]
      & \textbf{PW1} & 78 & 91 & 86 & 77 & \textbf{93} & \textbf{98} & 88 & 87.3 & 90.0 & 77.6\\[0.1cm]
      & \textbf{PW2} & 80 & 92 & 89 & 81 & \textbf{93} & \textbf{98} & 89 & 88.9 & \textbf{91.1} & 79.9 \\[0.1cm]
      & \textbf{PW3} & 81 & 93 & 89 & 82 & \textbf{93} & \textbf{98} & 89 & 89.5 & \textbf{91.4} & \textbf{80.5} \\[0.1cm]
      \midrule
      \parbox[t]{2mm}{\multirow{3}{*}{\rotatebox[origin=r]{90}{\textit{Parsing}}}} & \textbf{~\cite{teboul2011rl}} & 47 & 88 & 58 & 62 & 82 & 95 & 66 & 71.1 &  74.7 & -\\[0.1cm]
      & \textbf{\cite{martinovic2013bayesian}} & 50 & 81 & 49 & 66 & 80 & 91 & 71 & 69.7 & 74.8 & -\\[0.1cm]
      & \textbf{ST3+\cite{teboul2011rl}} & 64 & 90 & 71 & 76 & 93 & 96 & 77 & 81.0 & \textbf{85.2} & \textbf{72.4} \\
      \bottomrule
    \end{tabular}
    \caption{Segmentation results of various methods on
      ECP dataset. ST1, ST2, and ST3 correspond to the classification stages in our auto-context method. PW1,
      PW2, and PW3 refer to a Potts CRF model over the
      classification unaries. Published results are also shown for comparisons. The parsing
      results of the reinforcement learning method~\cite{teboul2011rl}
      when using the output ST3 result are reported in the last row.}
  \label{table:ecpresults}
\end{figure}

\subsection{Results on Single-view Segmentation}\label{sec:results}
The empirical results on different datasets are summarized in Fig.~\ref{table:ecpresults}
for the prominent ECP dataset and in Fig.~\ref{table:2dresults} for the remaining datasets,
where ST1, ST2, and ST3 correspond to the classification stages in
the auto-context method. In addition to the pixel-wise predictions of the auto-context
classifiers, we evaluated a CRF with an 8-connected neighbourhood and
pairwise Potts potentials. The single parameter of the Potts model
(weight for all classes set to equal) was optimized to yield the
highest accuracy on the training set (thus possibly at the expense of
losing a bit of performance compared to a cross-validation estimate).
Inference is done using alpha expansion implemented in DARWIN~\cite{Gould:JMLR2012}. The
results of the Potts-CRF on top of the unary predictions of
different staged auto-context classifiers are referred to as PW1, PW2, and PW3.

\begin{figure}[t]
  \scriptsize
  \centering
  \subfigure[eTRIMS dataset]{
    \begin{tabular}{p{1cm}>{\centering\arraybackslash}p{0.4cm}>{\centering\arraybackslash}p{0.4cm}>
        {\centering\arraybackslash}p{0.4cm}>{\centering\arraybackslash}p{0.5cm}>{\centering\arraybackslash}p{0.4cm}>
    {\centering\arraybackslash}p{0.4cm}>{\centering\arraybackslash}p{0.4cm}>{\centering\arraybackslash}p{0.4cm}}
      \toprule
      & \textbf{~\cite{martinovic2012three}} & \textbf{~\cite{cohen2014dp}} & \textbf{~\cite{gatta2014stacked}} & \textbf{\cite{kozinski2015mrfshapeprior}} & \textbf{ST1} & \textbf{ST2} & \textbf{ST3} & \textbf{PW3}\\ 
      \midrule
      \textbf{Average} & 65.3 & 65.9 & 66.4 & 66.0 & 74.0 & 76.8 & 76.7 &  \textbf{78.1}\\[0.1cm]
      \textbf{Overall} & 83.2 & 83.8 & 83.4 & 83.5 & 84.7 & 86.0 & 86.1 &  \textbf{87.3}\\[0.1cm]
      \textbf{\iou} & - & - & - & - & 58.7 & 61.3 & 61.5 & \textbf{63.5}\\
      \bottomrule
    \end{tabular}
  \label{table:etrimsresults}
  }

  \subfigure[CMP dataset]{
    \begin{tabular}{p{1.0cm}>{\centering\arraybackslash}p{0.7cm}>{\centering\arraybackslash}p{0.7cm}>{\centering\arraybackslash}p{0.7cm}>{\centering\arraybackslash}p{0.7cm}>{\centering\arraybackslash}p{0.7cm}}
      \toprule
      & \textbf{~\cite{Tylecek13}} & \textbf{ST1} & \textbf{ST2} & \textbf{ST3} & \textbf{PW3} \\
      \midrule
			\textbf{Average} & 47.5 & 40.5 & 47.0 & \textbf{48.7} & \textbf{48.9}\\[0.1cm]
			\textbf{Overall} & 60.3 & 61.8 & 65.5 & 66.2 & \textbf{68.1}\\[0.1cm]
			\textbf{\iou} & - & 29.3 & 34.5 & 35.9 & \textbf{37.5}\\
      \bottomrule
    \end{tabular}
  \label{table:cmpresults}
  }

  \subfigure[Graz dataset]{
    \begin{tabular}{p{1.0cm}p{0.7cm}>{\centering\arraybackslash}p{0.7cm}>{\centering\arraybackslash}p{0.7cm}>
        {\centering\arraybackslash}p{0.7cm}>{\centering\arraybackslash}p{0.7cm}>{\centering\arraybackslash}p{0.7cm}}
      \toprule
      & \textbf{~\cite{riemenschneider2012irregular}} & \textbf{~\cite{kozinski2015mrfshapeprior}} & \textbf{ST1} & \textbf{ST2} & \textbf{ST3} & \textbf{PW3}\\
      \midrule
      \textbf{Average} & 69 & \textbf{83.5} & 79.5 & 82.4 & 82.4 & \textbf{82.6}\\[0.1cm]
      \textbf{Overall} & 78 & \textbf{92.5} & 90.2 & 91.1 & 91.2 & \textbf{91.7}\\[0.1cm]
      \textbf{\iou}    & 58 & -             & 71.3 & 73.3 & 73.3 & \textbf{74.4}\\
      \bottomrule
    \end{tabular}
  \label{table:grazresults}
  }

  \subfigure[LabelMeFacades dataset]{
    \begin{tabular}{p{1cm}>{\centering\arraybackslash}p{0.7cm}>{\centering\arraybackslash}p{0.7cm}>{\centering\arraybackslash}p{0.7cm}>{\centering\arraybackslash}p{0.7cm}>{\centering\arraybackslash}p{0.7cm}>{\centering\arraybackslash}p{0.7cm}}
      \toprule
	  & \textbf{~\cite{frohlich2013semantic}} & \textbf{~\cite{nowozin2014optimal}} & \textbf{ST1} & \textbf{ST2} & \textbf{ST3} & \textbf{PW3}\\
      \midrule
			\textbf{Average} & \textbf{56.6} & - & 47.3 & 49.2 & 49.8  & 49.0		\\[0.1cm]
			\textbf{Overall} & 67.3 & 71.3 & 71.5 & 72.9 & 73.5  & \textbf{75.2}			\\[0.1cm]
			\textbf{\iou} & - & 36.0 & 37.0 & 38.7 & 39.4 & \textbf{39.6} \\
      \bottomrule
    \end{tabular}
  \label{table:labelmefacadesresults}
  }

  \subfigure[Art-deco dataset]{
    \begin{tabular}{p{1.0cm}>{\centering\arraybackslash}p{0.7cm}>{\centering\arraybackslash}p{0.7cm}>
        {\centering\arraybackslash}p{0.7cm}>{\centering\arraybackslash}p{0.7cm}>
        {\centering\arraybackslash}p{0.7cm}>{\centering\arraybackslash}p{0.7cm}}
      \toprule
      & \textbf{~\cite{gadde2016learninggrammars}} & \textbf{~\cite{kozinski2015mrfshapeprior} }& \textbf{ST1} & \textbf{ST2} & \textbf{ST3} & \textbf{PW3}\\ 
      \midrule
      \textbf{Average} & 72.9 & 83.8 & 80.8 & 84.0 & 84.3 & \textbf{84.8}\\[0.1cm]
      \textbf{Overall} & 78.0   & 88.8  & 85.9 & 88.1 & 88.3 & \textbf{89.0}\\[0.1cm]
         \textbf{\iou} & 58.0   & -     & 68.3 & 72.0 & 72.4 & \textbf{73.5}\\
      \bottomrule
    \end{tabular}
  \label{table:artdecoresults}
  }
  \caption{Segmentation results on various 2D datasets.
  ST1, ST2, and ST3 correspond to the classification stages in the auto-context method.
  And PW3 refer to a Potts CRF model over ST3 as unaries.}
  \label{table:2dresults}
\end{figure}

The first observation we make is that the use of a stacked
auto-context pipeline improves the results on \emph{all} the datasets. On the ECP
dataset, the improvement is 1.9\% in terms of overall pixel accuracy
for a three-stage classifier (ST3) compared to single-stage classifier
(ST1). The ordering in terms of statistically significant performance is
ST3$>$ST2$>$ST1 on the ECP, CMP, Art-deco and LabelMeFacade datasets and
ST3$=$ST2$>$ST1 on eTrims and Graz datasets. The auto-context features
are frequently selected
from the boosted decision trees. For the ECP dataset, about 30\% of the
features in stage 2 and 3 are auto-context features (CMP 46\%,
eTrims 31\%, and Graz 11\%). We didn't notice any significant differences or trends regarding the
type of auto-context features picked by the boosted decision trees for different datasets.

The next observation on the ECP dataset is that the overall accuracy of ST3,
with 90.8\%, is comparable with the reported 91.3\% from the current best
performing method~\cite{kozinski2015mrfshapeprior}. The CRF-Potts model (PW3) achieves
higher accuracies than the method of~\cite{kozinski2015mrfshapeprior}, making it the
(although only marginally) highest published result on the ECP
dataset. The results of the auto-context
classifier are significantly higher on the other datasets, except in the Graz dataset, when
compared to the methods
of~\cite{riemenschneider2012irregular,cohen2014dp,martinovic2012three,Tylecek13,frohlich2013semantic,nowozin2014optimal,gatta2014stacked,kozinski2015mrfshapeprior}.
On the eTRIMS, CMP and LabelMeFacades datasets, even the first stage classifier produces
better predictions than the previous approaches. The methods
of~\cite{riemenschneider2012irregular,cohen2014dp,martinovic2012three,Tylecek13,kozinski2015mrfshapeprior}
all include domain knowledge in their design. For example, the
system of~\cite{cohen2014dp} is a sequence of dynamic programs that 
include specific domain knowledge such as that balconies are
below windows, that only one door exists, or that elements like windows are
 rectangular segments.~\cite{kozinski2015mrfshapeprior} uses hand-written adjacency
patterns to limit the possible transitions between different states based on the semantic classes.
On the ECP dataset, the authors of~\cite{cohen2014dp} and~\cite{kozinski2015mrfshapeprior} observe
respectively, an improvement of about 4\% (personal communication) and 1.3\%
over their unary classifiers accuracy; we conjecture they also may improve the predictions of ST3.

The methods
of~\cite{riemenschneider2012irregular,cohen2014dp,martinovic2012three,Tylecek13,frohlich2013semantic,nowozin2014optimal}
use different unary predictions and therefore may profit from the
output of the auto-context classifier. Unfortunately, the respective
unary-only results are not reported, so at this point it is not
possible to estimate the relative improvement gains of the methods.
The fact that a conceptually simple auto-context pipeline
outperforms, or equals, all methods on all published datasets
suggests that a more careful evaluation of the relative improvements
of~\cite{riemenschneider2012irregular,cohen2014dp,martinovic2012three,Tylecek13,frohlich2013semantic}
is required.

\begin{figure}[t]
 \centering
 \scriptsize
 \setcounter{subfigure}{0}
     \centering
  \subfigure[ECP]{%
    \includegraphics[height=.11\textwidth]{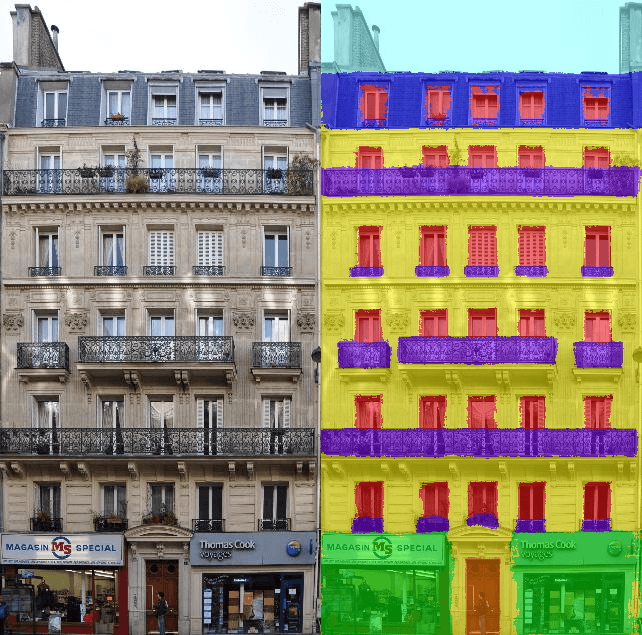} \label{fig:ecpsample}
  }
  \subfigure[eTRIMS]{%
    \includegraphics[height=.11\textwidth]{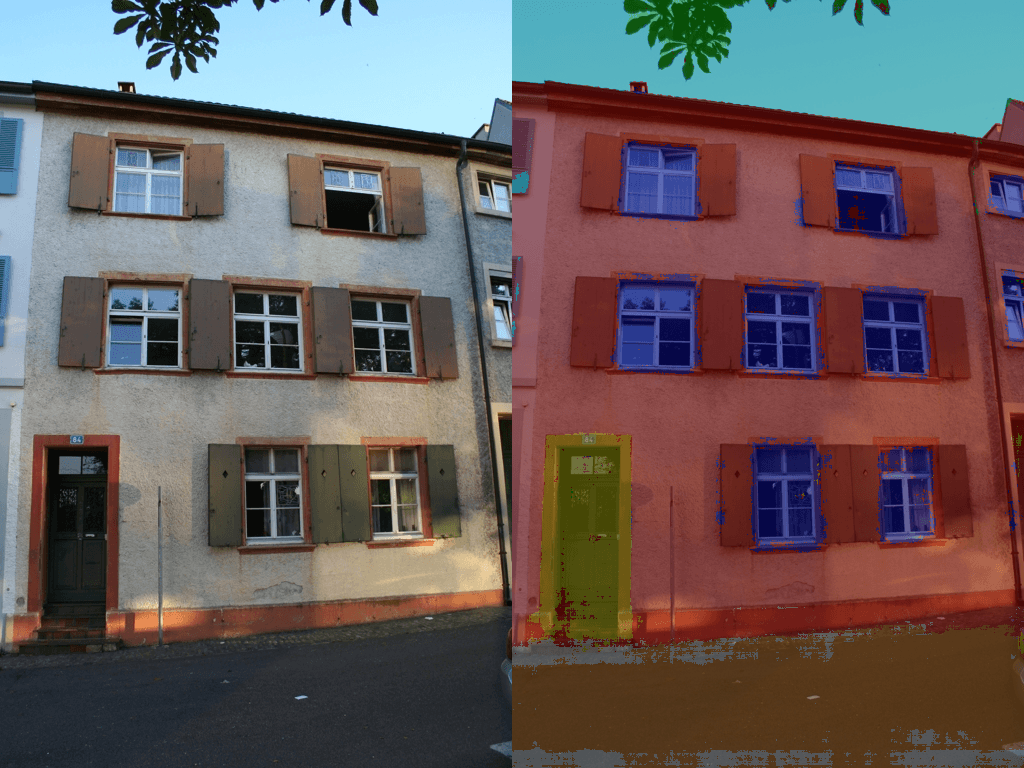} \label{fig:etrimssample}
  }
  \subfigure[Graz]{%
    \includegraphics[height=.11\textwidth]{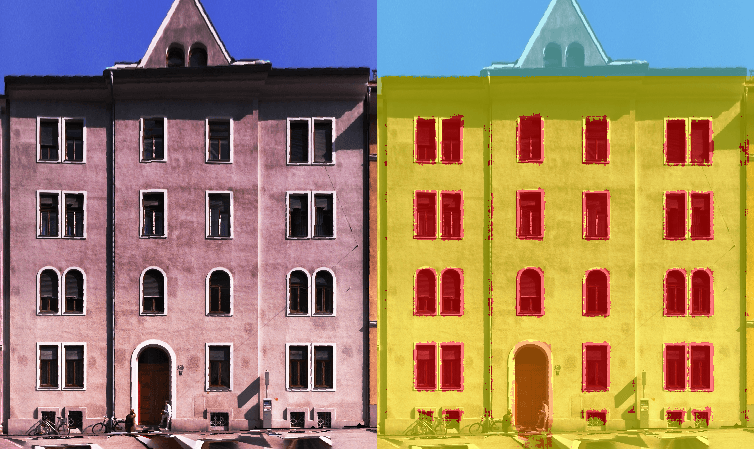} \label{fig:grazsample}
  }
  \subfigure[CMP]{%
    \includegraphics[height=.11\textwidth]{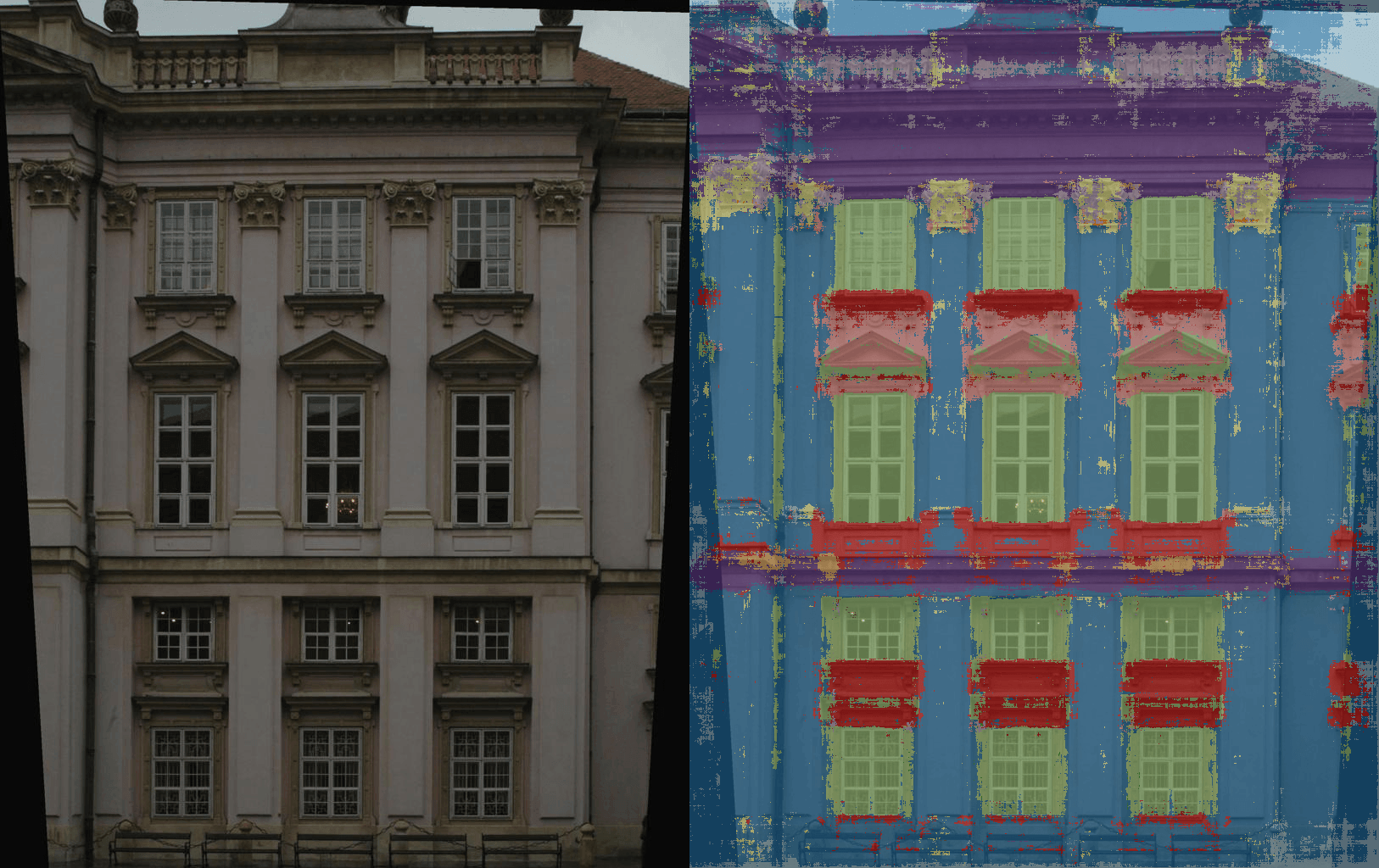} \label{fig:cmpsample}
  }
  \subfigure[ArtDeco]{%
    \includegraphics[height=.11\textwidth]{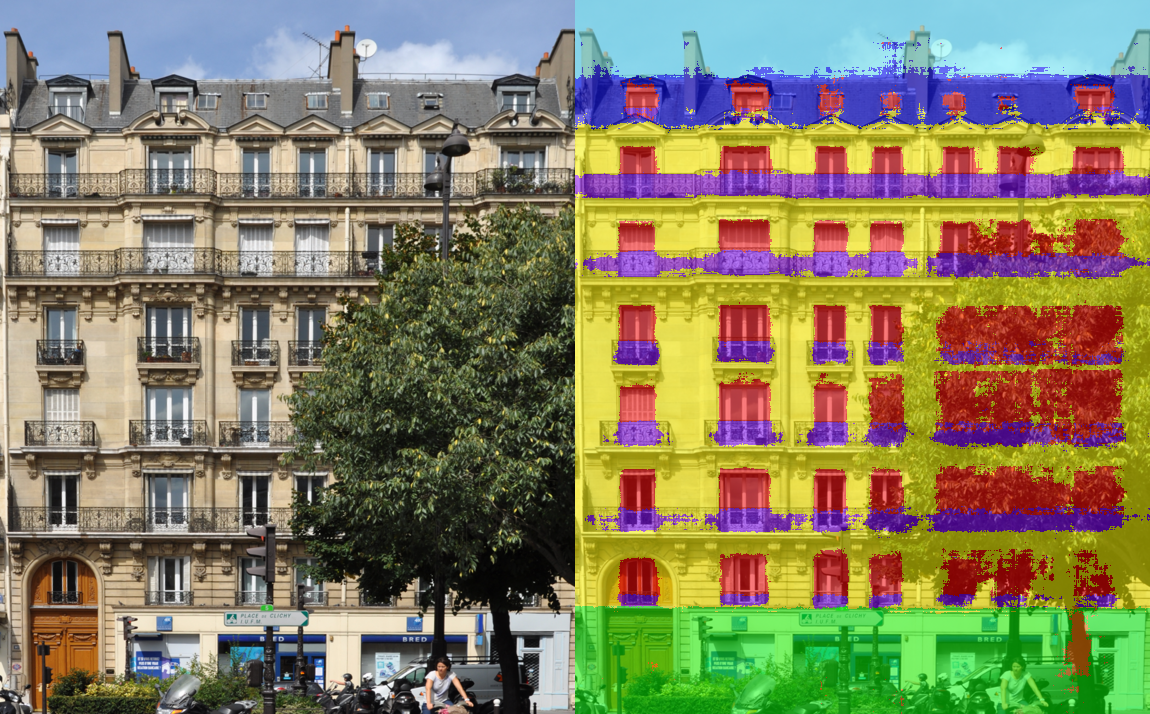} \label{fig:artdecosample}
  }
  \subfigure[RueMonge2014]{%
    \includegraphics[height=.11\textwidth]{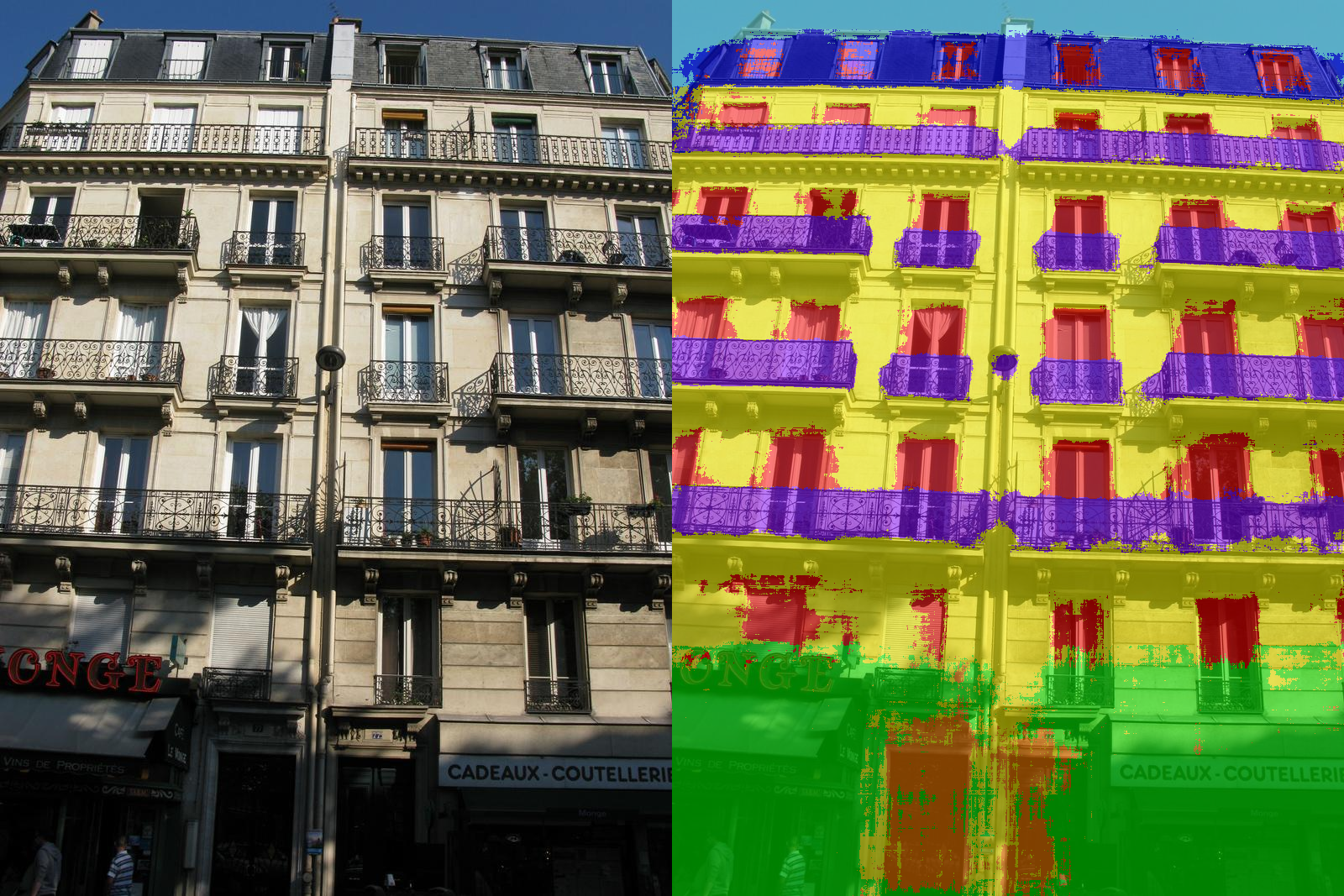} \label{fig:varcitysample}
  }
  \subfigure[LabelmeFacades]{%
    \includegraphics[height=.11\textwidth]{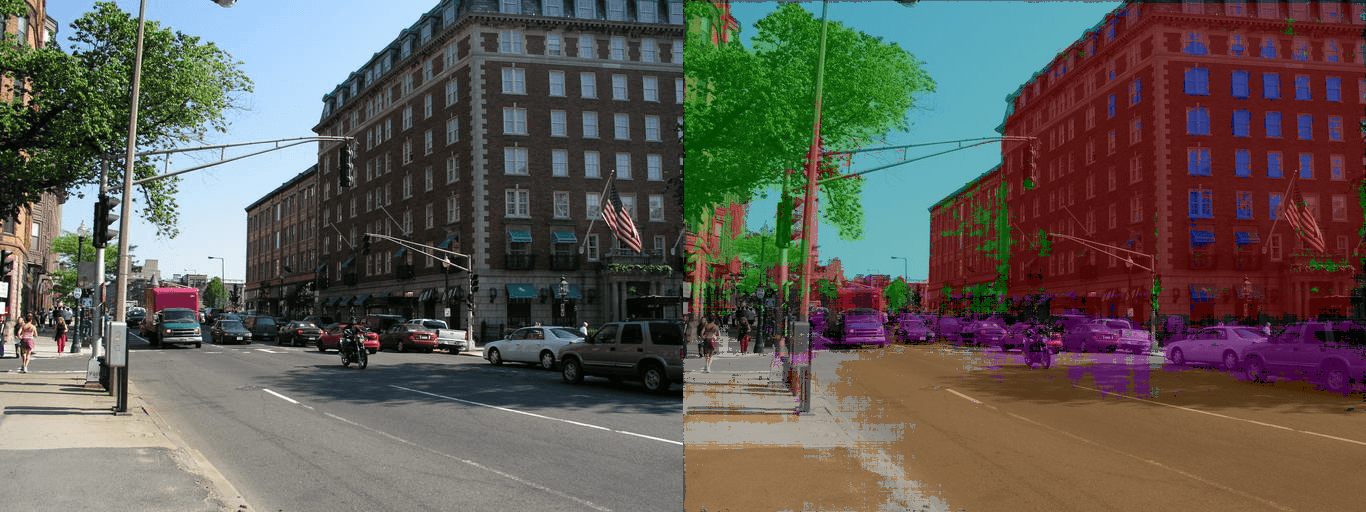} \label{fig:labelmefacadessample}
  }

  \caption{Sample segmentation visual results on different 2D facade datasets (best viewed in color and more in supplementary).}
  \label{fig:sample2dvisualresults}
\end{figure}

On all the datasets, we observe that the Potts model is
improving over the results from the auto-context stages ST1, ST2, and
ST3 ($p<0.01$). This suggests that some local statistics are not
captured in the auto-context features; more local features may improve
the auto-context classifiers. In practice, this performance gain has to be traded-off
against the inference time of alpha-expansion which is on average an
additional 24 seconds for an image from the ECP dataset. Some example visual results (ST3) are shown in
Fig.~\ref{fig:sample2dvisualresults} for different datasets and in Fig.~\ref{fig:sampleresult}
for different classification stages. All the results for comparison purpose are
publicly available at \url{http://fs.vjresearch.com}.

The average runtime of the system on 2D images
is summarized in Fig.~\ref{tbl:runtime}.
These numbers are computed on an Intel Core(TM) i7-4770 CPU @ 3.40 GHz
for a common image from the ECP dataset (about 500$\times$400 pixels).
All the extracted features are computed sequentially one after the other.
Fig.~\ref{tbl:runtime} also indicates that the proposed approach has favourable runtime
in comparison to existing prominent methods.

\begin{figure}[t]
  \scriptsize
  \setlength{\tabcolsep}{0pt}
  \centering
    \begin{tabular}{p{1.1cm}>{\centering\arraybackslash}p{0.9cm}>{\centering\arraybackslash}p{1.6cm}>{\centering\arraybackslash}p{0.7
    cm}>{\centering\arraybackslash}p{0.7cm}>{\centering\arraybackslash}p{0.7cm}>{\centering\arraybackslash}p{0.7cm}| >{\centering\arraybackslash}p{0.6cm}>{\centering\arraybackslash}p{0.6cm}>{\centering\arraybackslash}p{0.6cm}}
      \toprule
      \scriptsize
      \textbf{Method} & \textbf{Features} & \textbf{AC-Features} & \textbf{ST1} & \textbf{ST2} & \textbf{ST3} & \textbf{PW}  & \textbf{~\cite{martinovic2012three}} & \textbf{~\cite{cohen2014dp}} & \textbf{~\cite{kozinski2015mrfshapeprior}}\\ [0.2cm]
      \midrule
      \textbf{Time (s)} & 3.0 & 0.6 & +0.04 & +0.64 & +0.64 & +24 & +110 & +2.8 & +30\\
      \bottomrule
    \end{tabular}
    \caption{Average runtime for various methods on a single image of the ECP dataset.
    `Features' correspond to low-level and object detection image features (computed once).
    `AC-Features' corresponds to Auto-Context features.
    The classifier runs at 0.04 seconds, every stage needs to additionally compute AC features.
    A Potts model using alpha expansion takes on average 24s.
    Inference times (excluding unary computation) of existing methods are also shown for comparison.}
  \label{tbl:runtime}
\end{figure}

\begin{figure*}[t]
\setlength\abovecaptionskip{-1pt}
  \scriptsize
  \centering
  \subfigure[Image labeling]{
  \setlength{\tabcolsep}{0pt}
  \begin{tabular}{p{1.8cm}>
        {\centering\arraybackslash}p{1.0cm}>
        {\centering\arraybackslash}p{1.0cm}>{\centering\arraybackslash}p{1.0cm}>
        {\centering\arraybackslash}p{1.0cm}>{\centering\arraybackslash}p{1.0cm}|>
        {\centering\arraybackslash}p{1.0cm}>{\centering\arraybackslash}p{1.0cm}>
        {\centering\arraybackslash}p{1.0cm}>{\centering\arraybackslash}p{1.0cm}>{\centering\arraybackslash}p{1.0cm}|>
        {\centering\arraybackslash}p{1.0cm}>{\centering\arraybackslash}p{1.0cm}>{\centering\arraybackslash}p{1.0cm}>
        {\centering\arraybackslash}p{1.0cm}>{\centering\arraybackslash}p{1.0cm}>{\centering\arraybackslash}p{1.0cm}}
  \toprule
  & \multicolumn{5}{c}{\textbf{2D}} & \multicolumn{5}{c}{\textbf{3D}} &\multicolumn{6}{c}{\textbf{2D+3D}}\\
  \cmidrule(r){2-6}
  \cmidrule(r){7-11}
  \cmidrule(r){12-17}
  &\textbf{\cite{martinovic2012three}} & ST1 & ST2 & ST3 & PW3 & \textbf{\cite{riemenschneider2014learning}} & \textbf{\cite{martinovic20153d}}-3 & ST1 & ST2 & PW2 & \textbf{\cite{martinovic20153d}}-5 & \textbf{\cite{martinovic20153d}}-7 & ST3 & ST4 & PW3 & PW4\\
  \midrule
  \textbf{Average} & -    & 72.6& 72.8& 72.7& 73.7& -    & -    & 68.0 & 67.8&68.3&-    &-    &74.2&77.9&74.4&\textbf{79.0}\\[0.1cm]
  \textbf{Overall} & -    & 79.1& 79.4& 80.0& 81.2& -    & -    & 78.2& 82.0&82.3&-    &-    &\textbf{82.7}&80.9&83.4&81.9\\[0.1cm]
  \textbf{\iou}    & 57.5& 58.1& 58.4& 58.9& 60.5& 41.3& 53.2& 54.3& 56.3&57.0&61.3&62.0&62.0&61.2&\textbf{62.7}&62.7\\[0.1cm]
  \textbf{Runtime (min)} & 379  & 27   & 56   & 85   & 117  & 15   & 21   & 19   & 19   &20   &324  &404  &85   &114  &117  &146  \\
  \bottomrule
  \\
  \end{tabular}
  }
  \subfigure[Point Cloud labeling]{
  \setlength{\tabcolsep}{0pt}
  \begin{tabular}{p{1.8cm}>
        {\centering\arraybackslash}p{0.9cm}>{\centering\arraybackslash}p{0.9cm}>{\centering\arraybackslash}p{0.9cm}>
        {\centering\arraybackslash}p{0.9cm}>{\centering\arraybackslash}p{0.9cm}>
        {\centering\arraybackslash}p{0.9cm}>{\centering\arraybackslash}p{0.9cm}|>
        {\centering\arraybackslash}p{0.9cm}>
        {\centering\arraybackslash}p{0.9cm}>{\centering\arraybackslash}p{0.9cm}>
        {\centering\arraybackslash}p{0.9cm}>{\centering\arraybackslash}p{0.9cm}>{\centering\arraybackslash}p{0.9cm}|>
        {\centering\arraybackslash}p{0.9cm}>{\centering\arraybackslash}p{0.9cm}>
        {\centering\arraybackslash}p{0.9cm}>
        {\centering\arraybackslash}p{0.9cm}>{\centering\arraybackslash}p{0.9cm}}
  \toprule
  & \multicolumn{7}{c}{\textbf{2D}} & \multicolumn{6}{c}{\textbf{3D}} &\multicolumn{5}{c}{\textbf{2D+3D}}\\
  \cmidrule(r){2-8}
  \cmidrule(r){9-14}
  \cmidrule(r){15-19}
  &\textbf{\cite{martinovic2012three}} & \textbf{\cite{martinovic20153d}}-1 & \textbf{\cite{martinovic20153d}}-2 & ST1 & ST2 & ST3 & PW3 & \textbf{\cite{riemenschneider2014learning}} & \textbf{\cite{martinovic20153d}}-3 &  \textbf{\cite{martinovic20153d}}-4 & ST1 & ST2 & PW2 & \textbf{\cite{martinovic20153d}}-5 & \textbf{\cite{martinovic20153d}}-6 & ST3 & ST4 & PW4\\
  \midrule
  \textbf{Average} & -    & -    & -    & 70.8& 71.9& 72.6& 72.7& -    & -    & -    & 63.7& 68.0& 68.5 & -    & -    & 73.3 & \textbf{75.4} & 75.3\\[0.1cm]
  \textbf{Overall} & -    & -    & -    & 79.3& 80.9& 81.6& 82.1& -    & -    & -    & 78.8& 77.9& 78.6 & -    & -    & 84.3 & 84.4 & \textbf{84.7}\\[0.1cm]
  \textbf{\iou}    & 56.1& 55.7& 55.4& 55.7& 57.1& 58.2& 58.6& 42.3& 52.1& 52.2& 51.3& 53.6& 54.4 & 60.1& 60.8& 60.6 & 62.7 & \textbf{62.9}\\[0.1cm]
  \textbf{Runtime (min)} & 382  & 302  & 380  & 28   & 57   & 86   & 118  & 15   & 15   & 23   & 15   & 15   & 16    & 317  & 325  & 86    & 86     & 87\\
  \bottomrule
  \\
  \end{tabular}
  }

  \caption{Segmentation results of various methods for the tasks of (a) image labeling and (b) point cloud labeling on the RueMonge2014 dataset.
      In the following, L1 and L2 represent the first and second layers of~\cite{martinovic2012three}.
      RF refers to the random forest classifier in~\cite{martinovic20153d}.
      3DCRF refers to a Potts model-based CRF defined on 4-nearest neighbourhood of the point cloud.
      3DWR refers to weak architectural principles from~\cite{martinovic20153d}.
      The headings are defined as 
      ~\cite{martinovic20153d}-1 = L1+3DCRF,
      ~\cite{martinovic20153d}-2 = L1+3DCRF+3DWR,
      ~\cite{martinovic20153d}-3 = RF+3DCRF,
      ~\cite{martinovic20153d}-4 = RF+3DCRF+3DWR,
      ~\cite{martinovic20153d}-5 = RF+L1+3DCRF,
      ~\cite{martinovic20153d}-6 = RF+L1+3DCRF+3DWR, and
      ~\cite{martinovic20153d}-7 = RF+L2.
      The runtimes shown here, in minutes, include the feature extraction, classification and optional projection on the entire dataset.
      Note that, in case of 2D, the specified runtimes are the time taken to segment all 202 test images sequentially.
  }
  \label{table:varcityresults}
  \vspace{-0.3cm}
\end{figure*}

\subsection{Results on Multi-view Segmentation}\label{sec:results2}

In this section, we present semantic segmentation results of multi-view scenes using
2D images and 3D point clouds from the RueMonge2014 dataset~\cite{riemenschneider2014learning}.
The dataset details are already presented in Sec.~\ref{sec:datasets}.
Similar to~\cite{martinovic20153d}, we show results using only 2D images, only the 3D point cloud
and combined 2D+3D data for the tasks of image labeling and point cloud labeling.
Additionally, we present results for the mesh labeling task by projecting the image segmentation
results and point cloud segmentation results on to the mesh faces.

Here, we first apply the proposed auto-context technique, separately on 2D images and
3D point cloud to measure the relative gains. We find improvements in both 2D and 3D,
similar to the results on the single view datasets. For 2D images, the
performance levels out after three stages, while for 3D point clouds, the performance
levels out after two stages. This can be due to fewer auto-context features and less training
data for 3D point clouds. The publicly available
3D point cloud in the RueMonge2014 dataset has only 290196 training points and 276529
test points.

Next, we compare our results with the best performing techniques on these tasks.
All the results are summarized in Fig.~\ref{table:varcityresults}.
For all our experiments in 2D, we used the
specified training set of 119 images along with ground truth, and evaluated on
202 test images.
For the image labeling task, using only 2D images, we perform better (by +2.9\% \iou)
and faster (by at least a factor of 3) compared to~\cite{martinovic2012three}. Note that our
runtimes shown in Fig.~\ref{table:varcityresults} are computed by applying the proposed
technique sequentially on all the 202 test images. This can be easily parallelized by
performing the segmentation on all the test images in parallel.
Similar improvements in performance are observed in the image
labelling task when using only 3D point cloud data as well. Here, the image
labeling is performed by back-projecting the semantically-labeled 3D point cloud
onto 2D images.

For the point cloud labeling task, using either only 2D or 3D data,
we observe better segmentation results
compared to~\cite{martinovic20153d}.
We note that all improvements are obtained
without explicitly modeling structural priors or any other type of domain knowledge.
~\cite{martinovic20153d} proposed to use weak architectural principles in 3D on
top of initial segmentations that come from a simple classifier. Such weak
architectural principles have shown only a marginal improvement of +0.15\% in \iou.
In contrast, when using only 3D data,
the ST2 result of the proposed auto-context technique performs
better by +1.5\% in \iou, and improves even further by applying a pairwise Potts model.
Note that, in this case, we use exactly the same 3D features as~\cite{martinovic20153d}
for the first stage classification. This renders the results of weak architectural
principles in~\cite{martinovic20153d} and the proposed auto-context features directly comparable.
We also obtain favorable runtimes. It takes 8 minutes to enforce the weak achitectural principles but
less than a minute to apply another stage of auto-context.

\begin{figure}[ht]
  \scriptsize
  \centering
  \begin{tabular}{p{1.6cm}>
        {\centering\arraybackslash}p{0.5cm}>{\centering\arraybackslash}p{0.35cm}>
        {\centering\arraybackslash}p{0.35cm}>
        {\centering\arraybackslash}p{0.35cm}>{\centering\arraybackslash}p{0.35cm}>
        {\centering\arraybackslash}p{0.35cm}>
        {\centering\arraybackslash}p{0.35cm}>{\centering\arraybackslash}p{0.35cm}}
  \toprule
  & \multirow{2}{*}{\textbf{\cite{riemenschneider2014learning}}}& \multicolumn{4}{c}{\textbf{2D}} &\multicolumn{3}{c}{\textbf{2D+3D}}\\
  \cmidrule(r){3-6}
  \cmidrule(r){7-9}
  & & ST1 & ST2 & ST3 & PW3 & ST3 & ST4 & PW4\\
  \midrule
  \textbf{Average} &   - &71.6&72.5&72.2&72.2&73.2&79.9&\textbf{80.0}\\[0.1cm]
  \textbf{Overall} &   - &81.7&82.3&82.5&82.8&84.4&84.1&\textbf{84.4}\\[0.1cm]
     \textbf{\iou} & 41.9&57.8&59.0&58.7&58.8&60.9&63.2&\textbf{63.7}\\[0.1cm]
  \textbf{Runtime (min)} & 15   &31   &60   &89   &121  &89   &118  &150\\
  \bottomrule
  \end{tabular}
  \caption{Results for mesh labeling task on the RueMonge2014 dataset.}
  \label{table:varcitymeshresults}
\end{figure}

\begin{figure*}[t]
\vspace{-0.5cm}
\setlength\abovecaptionskip{-1pt}
\centering
\subfigure[Point Cloud]{\includegraphics[height=.13\textwidth]{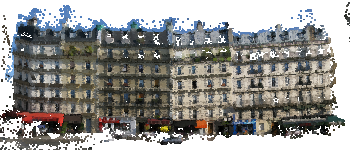}%
\label{fig_first_case}}
\subfigure[Ground Truth]{\includegraphics[height=.13\textwidth]{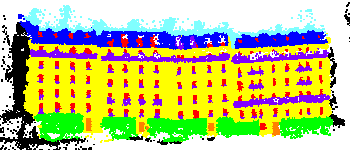}%
\label{fig_second_case}}
\subfigure[ST4]{\includegraphics[height=.13\textwidth]{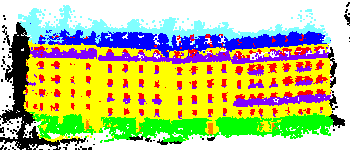}%
\label{fig_second_case}}
\vspace{-0.2cm}

\centering
 \setcounter{subfigure}{0}
\subfigure[Mesh Surface]{\includegraphics[height=.14\textwidth]{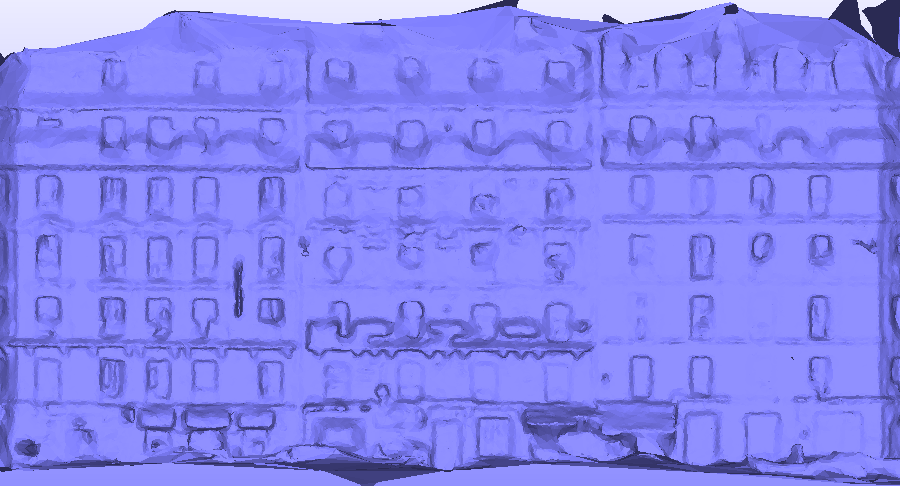}%
\label{fig_first_case}}~~~~~~~~
\subfigure[Ground Truth]{\includegraphics[height=.14\textwidth]{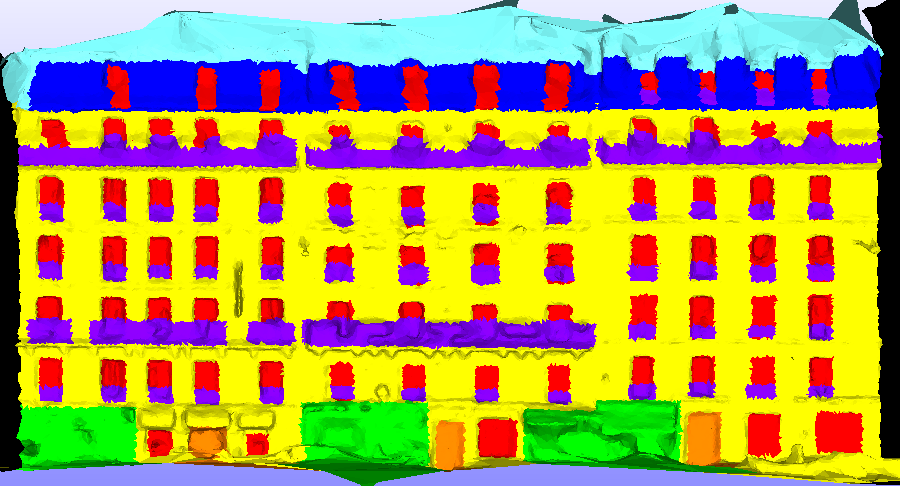}%
\label{fig_second_case}}~~~~~~~~
\subfigure[ST4]{\includegraphics[height=.14\textwidth]{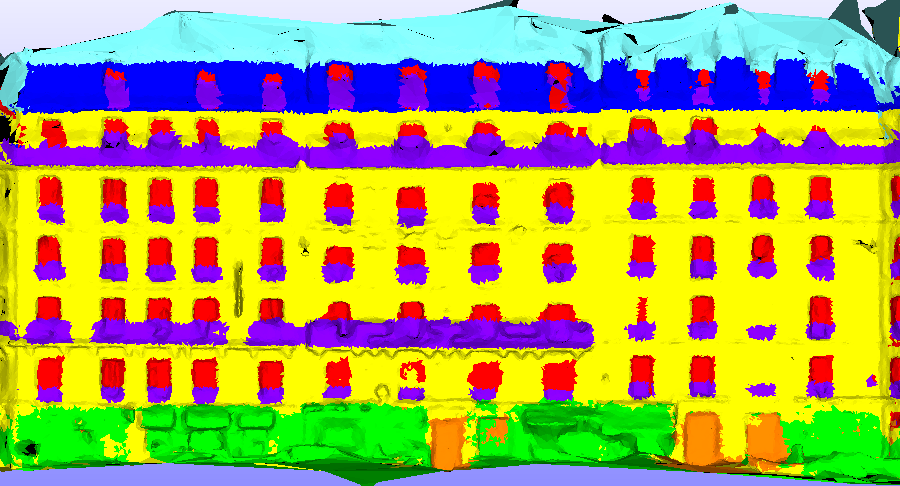}%
\label{fig_second_case}}
\caption{Sample visual results for the point cloud and  mesh labeling tasks on the RueMonge2014 dataset (More in supplementary).}
\label{fig_3d_visuals}
\end{figure*}

\begin{figure*}[ht]
\setlength\abovecaptionskip{-1pt}
\vspace{-0.2cm}
\label{fig:sample}
 \centering

    \setcounter{subfigure}{0}
     \begin{minipage}[c]{\textwidth}
     ~~~~    \subfigure[Facade]{%
    \includegraphics[width=.105\textwidth]{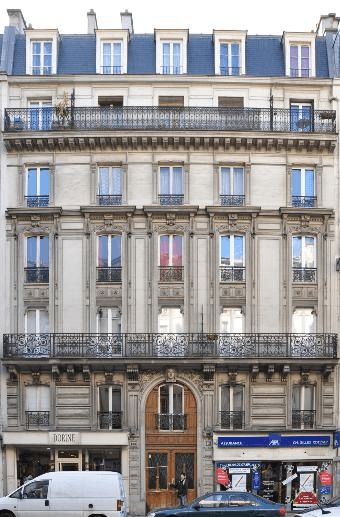} \label{fig:sampleimage}
  }~~~~
  \subfigure[GT]{%
    \includegraphics[width=.105\textwidth]{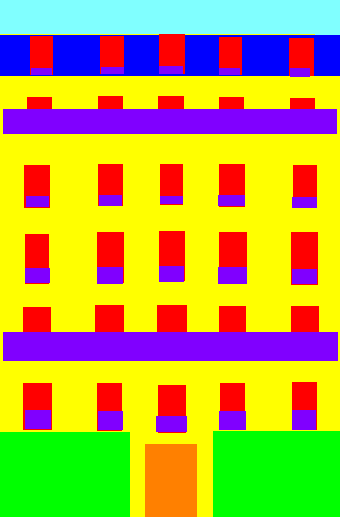} \label{fig:groundtruth}
  }~~~~
  \subfigure[ST1]{%
    \includegraphics[width=.105\textwidth]{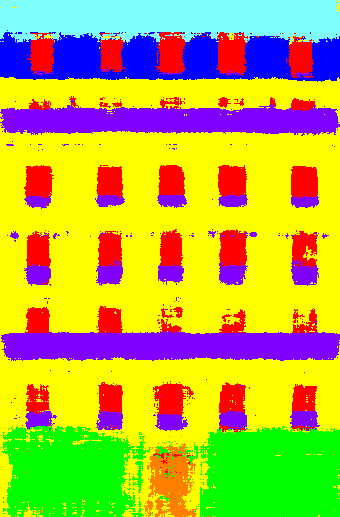} \label{fig:stage1}
  }~~~~
  \subfigure[ST2]{%
    \includegraphics[width=.105\textwidth]{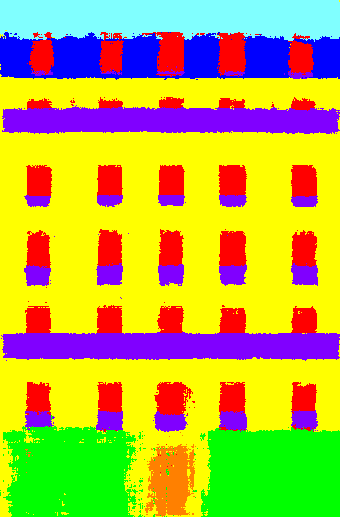} \label{fig:stage2}
  }~~~~
  \subfigure[ST3]{%
    \includegraphics[width=.105\textwidth]{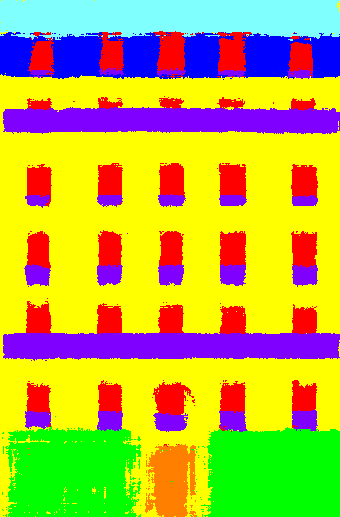} \label{fig:stage3}
  }~~~~
  \subfigure[PW3]{%
    \includegraphics[width=.105\textwidth]{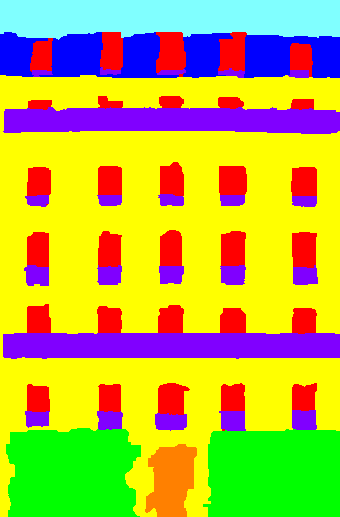} \label{fig:pw3}
  }~~~~
  \subfigure[Parse]{%
    \includegraphics[width=.105\textwidth]{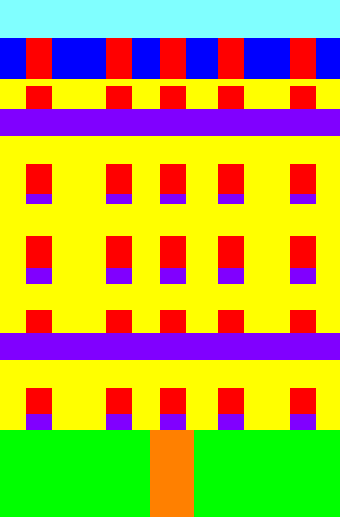} \label{fig:rlresult}
  }
  \end{minipage}

  \caption{(a) Sample facade image from ECP dataset; (b) Ground truth
    segmentation; and (c,d,e) Result of various classification stages
    of our auto-context method. Observe that the method removes
    isolated predictions and recovers the second lowest line of
    windows. (f) Potts model on top of ST3 result, and (g) parsed
    result obtained by applying reinforcement
    learning~\cite{teboul2011rl} using ST3 result.}
    \label{fig:sampleresult}
\vspace{-0.4cm}
\end{figure*}

Next, similar to~\cite{martinovic20153d}, we combine the segmentation results of 2D images
and 3D point clouds for further improving the performance (`2D+3D' in Fig.~\ref{table:varcityresults}).
For this, we accumulate the ST3
and ST2 results of 2D images and 3D point cloud respectively. Further applying auto-context (ST4) on the
accumulated unaries boosts the \iou~performance by another 2.1\% in labeling the point cloud, while it
levels out in labeling the images. Additionally, applying a pairwise Potts CRF model similar
to~\cite{martinovic20153d} to enforce smoothness, further increases the
\iou~performance by 0.7\% and 0.2\% in labeling the images and point cloud respectively.
In summary, as evident in Fig.~\ref{table:varcityresults}, better performance than existing state-of-the-art
approaches is obtained in both image and 3D point cloud labeling,
while being 2-3 times faster. A visual result
of 3D point cloud segmentation is shown in Fig.~\ref{fig_3d_visuals}

Finally, we compare our results with existing approaches on the mesh labeling problem.
To label meshes, the 2D image segmentation results are projected onto the mesh faces.
2D segmentation results obtained with only 2D images and with both 2D+3D data are used for
mesh labeling. See Fig.~\ref{table:varcitymeshresults} for quantitative results and Fig.~\ref{fig_3d_visuals} for
a visual result. Again we observe similar improvements with auto-context classifiers.
Our simple majority voting scheme to project the semantics from 2D images to 3D mesh shows
that we perform significantly better (by +17.4\% IoU) in comparison to the existing
approach from~\cite{riemenschneider2014learning}.
However this comes at a cost of runtime (6 $\times$ slower). The result of earlier stages of our
auto-context technique still performs better, `ST1' for example, by +15.9\% with a comparable
runtime. In \cite{riemenschneider2014learning}, 3D information is used to reduce redundant evaluations,
thereby achieving faster runtime.

\vspace{-0.2cm}
\subsection{Inverse Procedural Modeling}
\label{sec:modeling}
A pixel-wise classification of a facade might not be the
desired input for some applications. This fact motivated shape grammar
methods~\cite{riemenschneider2012irregular,ripperda06rjmcmc,teboul2011rl,martinovic2013bayesian}
that parse the facade into a high-level structured representation.
The aim of these top-down approaches is to infer the
architectural (structural) information in facades by fitting a set of
grammar rules (a derivation) to a pixel classifier output. Such structural information can
be used for 
retrieving structurally similar facades, etc. We apply the parsing method
of~\cite{teboul2011rl} and compare against their result, that is
obtained using a random forest classifier that uses color information.
All other settings and the grammar are the same. We
refer the reader to~\cite{teboul2011rl} for more details about the approach. The
results are shown in the last three rows of
Fig.~\ref{table:ecpresults}. These numbers are obtained by
back-projecting the parsed representation into a pixel-wise
prediction. We observe that better pixel predictions directly
translates to better parsing results. A substantial improvement of
10.5\% is achieved, closing the gap to pixel prediction methods. This
shows the importance of good pixel predictions even for models that
only make use of them as an intermediate step. Fig.~\ref{fig:sampleresult}
shows a sample visual result of various classification stages and
the parsing result obtained with ST3 + \cite{teboul2011rl}.

\vspace{-0.2cm}
\section{Conclusion}
\label{sec:conclusion}

The segmentation method that we described in this paper is a framework
of established and proven components. It is easy to implement, fast at test-time,
and it outperforms all previous approaches on all published facade
segmentation datasets. It is also the fastest method amongst all those that
we compared against. The runtime is dominated by feature computation,
which is amenable to massive speed improvements using parallelization in case a high-performing
implementation is required.

We observe on all datasets that
adding stacked classifiers using auto-context features improves the performance.
This applies to both 2D (images) and 3D (point clouds) data.
For the ECP dataset, a Potts-CRF further improves the performance but this comes at the expense of a severe increase in runtime.
Further, the proposed technique can be applied independently to either
2D or 3D data and also to combined 2D+3D models. For the point cloud labeling
task, on RueMonge2014 dataset, applying auto-context on the combined 2D+3D
improves the \iou~performance by 1.9\%.

The proposed auto-context classifier raises the bar when it comes to absolute
performance. Contrary to the popular belief in this domain,
it largely ignores domain knowledge, but still 
performs better than all the methods that include prior
information in some form, for example
relationship between balconies and windows. We believe that it is
important to evaluate methods in terms of a relative improvement over
strong pixel classifier baselines. In order to facilitate a fair
comparison of previous and future work, we release all code
that has been used to obtain the reported
results along with all predictions for easy comparison\footnote{\url{http://fs.vjresearch.com}}.

\ifCLASSOPTIONcaptionsoff
  \newpage
\fi

\small\section*{Acknowledgements}
This work was partly carried out in IMAGINE, a joint research project between ENPC and CSTB, and partly supported by ANR-13-CORD-0003 and ECP.
\normalsize

\small
{
\bibliographystyle{IEEEtran}
\bibliography{paper}
}

\end{document}